\documentclass[hidelinks,letter, 10 pt, conference,doublecolomn]{ieeeconf} 
\IEEEoverridecommandlockouts    
\usepackage{balance}

\usepackage{mathtools}
\usepackage{mathrsfs} \usepackage{bm}
\usepackage{subfiles}
\usepackage{graphicx}
\usepackage{epstopdf}
\usepackage{float}

\usepackage{bm}
\usepackage{amsmath}
\usepackage{amssymb}

\newcommand{\vect}[1]{\boldsymbol{#1}}
\newcommand{\mat}[1]{\boldsymbol{#1}}

\newcommand{\diffs}[3]{\frac{\partial^2 #1}{
\ifx#2#3 
\partial #2^2
\else
\partial #2 \partial #3
\fi
}}

\newcommand{\av}{\vect{a}}

\newcommand{\dv}{\vect{d}}
\newcommand{\ev}{\vect{e}}

\newcommand{\fv}{\vect{f}}

\newcommand{\hv}{\vect{h}}

\newcommand{\mv}{\vect{m}}
\newcommand{\nv}{\vect{n}}

\newcommand{\pv}{\vect{p}}

\newcommand{\rv}{{\vect{r}}}

\newcommand{\uv}{\vect{u}}
\newcommand{\vv}{\vect{v}}

\newcommand{\wv}{\vect{w}}

\newcommand{\xv}{\vect{x}}

\newcommand{\yv}{\vect{y}}

\newcommand{\zv}{\vect{z}}

\newcommand{\muv}{\bm{\mu}}

\newcommand{\Omegav}{\bm{\Omega}}

\newcommand{\nuv}{\vect{\nu}}
\newcommand{\omegav}{\bm{\omega}}

\newcommand{\Am}{\mat{A}}

\newcommand{\Dm}{\mat{D}}

\newcommand{\Fm}{\mat{F}}
\newcommand{\Gm}{\mat{G}}

\newcommand{\Jm}{\mat{J}}
\newcommand{\Lm}{\mat{L}}

\newcommand{\Km}{\mat{K}}

\newcommand{\Rm}{\mat{R}}

\newcommand{\Sigmam}{\bm{\Sigma}}

\usepackage{svg}
\usepackage{hyperref}
\usepackage{academicons}
\usepackage{xcolor}

\usepackage{amsthm}
\theoremstyle{plain}

\usepackage{comment}
\usepackage[font=small]{caption}
\usepackage{subcaption}

\usepackage{calc}
\newtheorem{rem}{\textbf{Remark}}[section]
\usepackage{mathtools} 
\title{\LARGE \bf
A Comparative Study of INDI and NDI with Nonlinear Disturbance Observer for Aerial Robotics
}
\author{Benedetta Rota$^3$, Mirko Mizzoni$^1$\orcidlink{0009-0006-2165-3475}, Amr Afifi$^1$\orcidlink{0000-0002-2267-575X},  Pieter van Goor{$^2$\orcidlink{0000-0003-4391-7014}}, and Antonio Franchi$^{1,3}$\orcidlink{0000-0002-5670-1282} 
\thanks{$^1$Robotics and Mechatronics group, Faculty of Electrical Engineering,  Mathematics, and Computer Science (EEMCS), University of Twente, 7500 AE Enschede, The Netherlands. {\footnotesize \tt m.mizzoni@utwente.nl,} {\footnotesize \tt amrafifi142@gmail.com,}{\footnotesize \tt schol@r-franchi.eu}}
\thanks{$^{2}$ School of Aerospace, Mechanical, and Mechatronic Engineering (AMME), Faculty of Engineering, University of Sydney, NSW, 2006, Australia. {\footnotesize \tt pieter.vangoor@sydney.edu.au}}
\thanks{$^3$Department of Computer, Control and Management Engineering, Sapienza University of Rome, 00185 Rome, Italy, {\footnotesize \tt schol@r-franchi.eu}}
\thanks{Accepted for presentation at the 2026 International Conference on Unmanned Aircraft Systems (ICUAS 2026).}
\thanks{This work was partially funded by the Horizon Europe research projects 101120732 (AUTOASSESS) and MSCA PF  MEW (101154194)}}

\usepackage{orcidlink}

\let\temp\rmdefault
\usepackage{mathpazo}
\let\rmdefault\temp

\newif\ifappendix
\appendixfalse 

\begin{document}
\maketitle

\begin{abstract}
This work presents  a simulation-based comparative robustness analysis of
Incremental Nonlinear Dynamic Inversion (INDI) 
and Nonlinear Dynamic Inversion augmented with a nonlinear disturbance observer (NDI+NDO) for fully actuated aerial robots. A systematic simulation campaign across representative operating scenarios is conducted, where we compare tracking performance, robustness,  control effort, under parametric variations, external disturbances, and measurement noise. Results show that  INDI demonstrates stronger robustness in several model-mismatch and combined-stress cases, while NDI+NDO primarily matches nominal performance but exhibits greater sensitivity under several non-ideal conditions. These findings provide practical guidance on the relative strengths and limitations of incremental and observer-based inversion strategies for aerial robotic applications.
\end{abstract}
\section{Introduction}
Aerial robotics platforms have become fundamental in modern engineering applications, ranging from environmental monitoring and precision agriculture to infrastructure inspection, search-and-rescue operations, and physical interaction tasks~\cite{2023e-AfiCorSabAboSidFra,colomina2014unmanned}. The rapid development of multirotor unmanned aerial vehicles (UAVs) has been enabled by advances in sensing, embedded computation, and control design, allowing these platforms to operate autonomously in increasingly complex and uncertain environments. However, the performance and safety of aerial robots critically depend on the robustness of their control systems, particularly in the presence of aerodynamic disturbances, parametric uncertainties, and sensor noise.

The dynamics of multirotor platforms are inherently nonlinear and strongly coupled. As a consequence, nonlinear control techniques have become standard tools in flight control design. Among these, Nonlinear Dynamic Inversion (NDI) has established itself as a widely adopted methodology in aerospace applications due to its systematic input–output linearization capability and clear stability properties,~\cite{Isidori1995}. By explicitly canceling nonlinearities through model-based feedback, NDI enables the designer to impose desired linear tracking dynamics on the closed-loop system.

\begin{figure}[t!]
        \centering \includegraphics[width=1.02\linewidth]{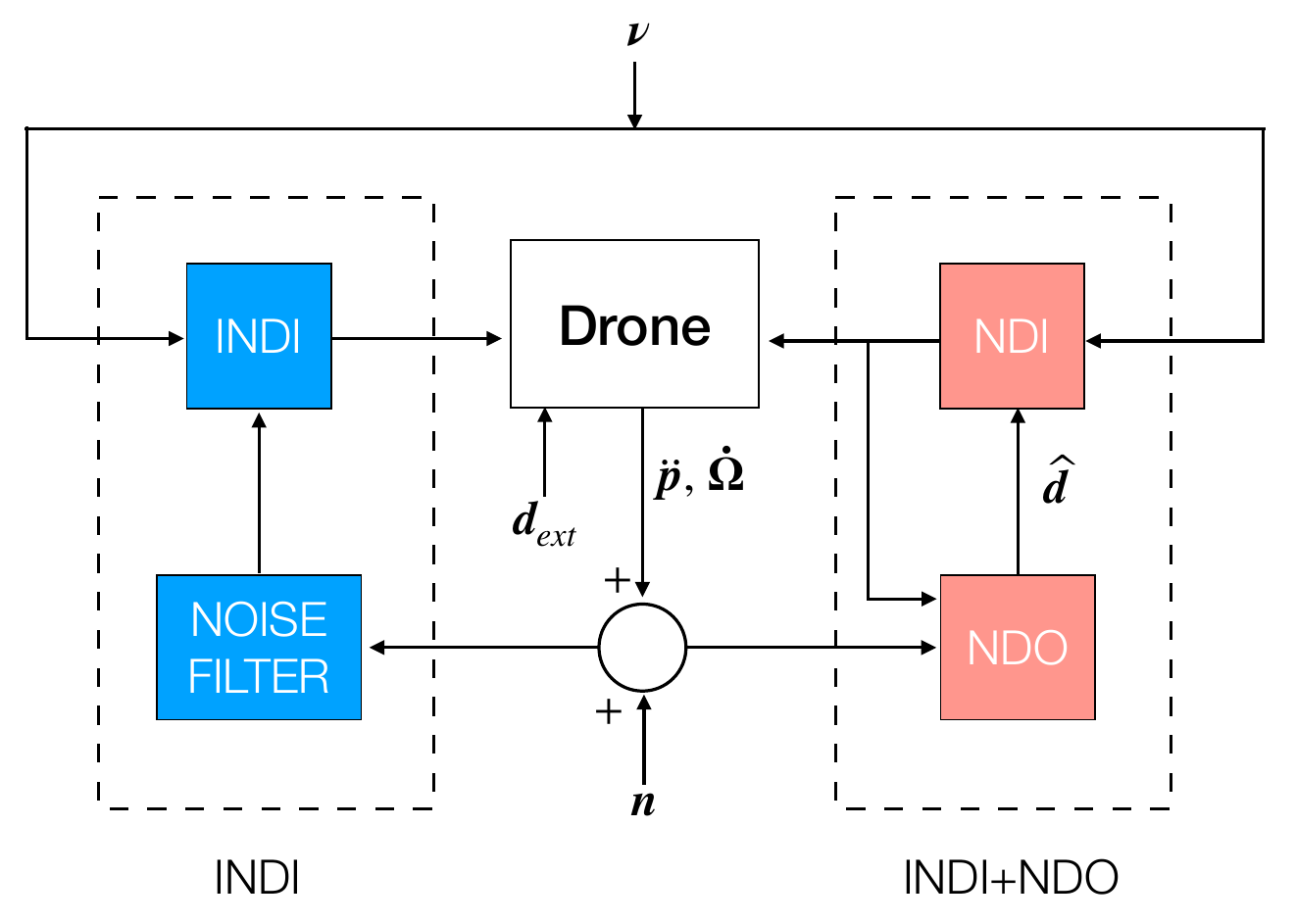}
    \caption{Block-diagram comparison between the INDI and NDI+NDO control architectures. Both schemes regulate the same robotic platform under the same reference signal $\nuv$, external disturbances $\dv_{\mathrm{ext}}$, and measurement noise $\nv$. INDI computes control updates through incremental feedback based on filtered measured response, whereas NDI+NDO incorporates explicit disturbance estimation through a nonlinear observer within the inversion loop.}
    \label{fig:placeholder}
\end{figure}
Despite its theoretical appeal, the practical performance of NDI is highly sensitive to model accuracy. In aerial vehicles, parameters such as mass and inertia may vary due to payload changes, while aerodynamic forces and moments are often difficult to model precisely. External disturbances, such as wind gusts, further challenge exact model cancellation. In such situations, modeling errors directly affect the inversion process, potentially leading to degraded tracking performance or increased control effort. To mitigate these limitations, several extensions have been proposed, including robust outer-loop designs, adaptive schemes, and disturbance observer-based augmentation. While effective, these solutions increase architectural complexity and tuning burden.

In this context, Incremental Nonlinear Dynamic Inversion (INDI) has emerged as a practical alternative, particularly in the field of multirotor UAVs and micro aerial vehicles,~\cite{Sun2021,liu_2022,wang2019,Sieberling2010}. Rather than recomputing the full nonlinear cancellation at every control step, INDI exploits measured accelerations or output derivatives to compute incremental control updates. By relying on short-term variations in the system response, INDI implicitly incorporates the effects of modeling errors and disturbances without requiring explicit disturbance models. This reduced dependence on the nominal model has made such control technique attractive for agile flight, disturbance rejection, and real-world implementation on lightweight aerial platforms,~\cite{acquatella2012}.


However, in the current literature, a systematic comparative analysis between INDI and a classic NDI architecture augmented with a nonlinear disturbance observer (NDI+NDO) is still missing.
Nonlinear disturbance observers based on residual formulations have been proposed in the aerial robotics literature, notably the contact force estimation scheme presented in~\cite{ryll}.
This motivates the following practical question:
\begin{quote}
{
``How do INDI and NDI+NDO compare in terms of robustness, tracking performance, and control effort across nominal and non-ideal operating conditions?''}
\end{quote}

In nominal disturbance-free conditions, the considered INDI and NDI+NDO formulations  produce identical control inputs for the fully actuated platform studied in this work, providing an analytical baseline for comparison. However, this nominal correspondence does not necessarily extend to practical implementations, where discretization, filtering, model uncertainty, external disturbances, and measurement noise introduce meaningful differences in behavior.

This work addresses this comparison through an analytical nominal-condition study together with a systematic simulation campaign under realistic operating scenarios, including wind disturbances, parametric uncertainty, energy consumption, and sensing degradation. The analysis is conducted on a dynamically demanding Lissajous trajectory to evaluate tracking performance, robustness margins, and control effort across progressively challenging conditions. The results show that, across the considered simulation campaign, INDI consistently matches or outperforms NDI+NDO, with the largest advantages emerging under model mismatch, sensing degradation, and combined stress conditions. This provides practical insight into the robustness and control-effort benefits of incremental inversion for the considered aerial robotic platform and operating scenarios.


The remainder of the paper is organized as follows. 
Section~\ref{sec:equivalence} presents an analytical comparison under nominal conditions.
Section~\ref{sect:sys_modeling} presents the multi-rotor dynamic model and the control architectures. 
Section~\ref{sec:robustness_framework} describes the robustness campaign setup and performance metrics. 
Section~\ref{sec:discussion} discusses simulation results and comparative analysis. Finally, Section~\ref{sec:concl} concludes the paper and outlines future research directions.


\section{Analytical Comparison in Nominal Conditions}
\label{sec:equivalence}

Consider the control-affine nonlinear system
\begin{equation}
\dot{\xv} = \fv(\xv) + \Gm(\xv)\uv + \wv, 
\qquad
\yv = \hv(\xv),
\end{equation}
where $\xv \in \mathbb{R}^n$ is the state, $\uv \in \mathbb{R}^m$ is the control input,  \mbox{$\yv \in \mathbb{R}^p$} is the output vector, $\wv \in \mathbb{R}^n$ is an unknown disturbance, and $\fv,\Gm,\hv$ are assumed to be sufficiently smooth (at least $C^2$) on a domain $U \subset \mathbb{R}^n$.

We assume w.l.o.g. that the system has  a well-defined (vector) relative degree one 
with respect to each entry of the output $\yv$ on $U$, i.e.,
$
\Am(\xv):=\frac{\partial \hv}{\partial \xv}(\xv)\Gm(\xv)
$
has full rank for all $\xv \in U$. Then, the input–output 
dynamics can be written as
\begin{equation}
\dot{\yv} = \av(\xv) + \Am(\xv)\uv + \dv,
\label{eq:io}
\end{equation}
where $
\av(\xv) := L_{\fv}\hv(\xv),$ 
$\dv := L_{\wv}\hv(\xv)$,  with the Lie derivatives defined as
$
L_{\fv}\hv(\xv) := \frac{\partial \hv}{\partial \xv}(\xv)\fv(\xv), 
$
and
$
L_{\wv}\hv(\xv) := \frac{\partial \hv}{\partial \xv}(\xv)\wv(\xv,t)
$.
The matrix $\Am(\xv)$ is commonly referred to as the decoupling matrix~\cite{Isidori1995}. 
We restrict attention to the square case $m=p$ and assume that 
$\Am(\xv)$ is invertible for all $\xv \in U$.
The derivation below is intended to clarify input-level relationships under nominal assumptions and provide analytical motivation for comparison.   We assume that the control sampling period $T_s$ is significantly smaller than the characteristic time constant of the closed-loop system, such that the control input is approximately constant over one update interval, i.e., $\uv(t) \approx \uv_f:=\uv(t - T_s)$.

Given a virtual input $\nuv\in \mathbb{R}^p$, the control objective is to design $\uv$ such that $
\dot{\yv} = \nuv$.
NDI augmented with a nonlinear disturbance observer (NDO) is given by
\begin{equation}
\uv_{\mathrm{NDO}} = \Am(\xv)^{-1}(\nuv - \av(\xv) - \widehat{\dv}),
\label{eq:ndo_control}
\end{equation}
with observer dynamics inspired by~\cite{ryll},
\begin{equation}\label{eq:ob_dyn}
\dot{\widehat{\dv}} = -\Lm\widehat{\dv} + \Lm(\dot{\yv} - \av(\xv) - \Am(\xv)\uv),
\end{equation}
where $\Lm>0$. 
INDI instead computes
\begin{equation}
\uv_{\mathrm{INDI}} = \uv_f + \Am(\xv)^{-1}(\nuv - \dot{\yv}_f),
\label{eq:indi}
\end{equation}
where $\dot{\yv}_f = \dot{\yv} + \varepsilon_f$ denotes the filtered derivative of the output, and $\uv_f$ is the previously applied input.

Define the incremental disturbance estimate
\begin{equation}
\widehat{\rv} = \dot{\yv}_f - \av(\xv) - \Am(\xv)\uv_f.
\end{equation}

Adding and subtracting $\dot{\yv}_f - \Am(\xv)\uv_f$ inside \eqref{eq:ndo_control} yields
\begin{align}
\uv_{\mathrm{NDO}}
&= \Am(\xv)^{-1}\Bigl(\nuv - \av(\xv) - \widehat{\dv}
+ \dot{\yv}_f - \Am(\xv)\uv_f \Bigr) \nonumber \\
&\quad + \Am(\xv)^{-1}\Bigl(- \dot{\yv}_f + \Am(\xv)\uv_f\Bigr) \nonumber \\
&= \uv_f + \Am(\xv)^{-1}(\nuv - \dot{\yv}_f)
+ \Am(\xv)^{-1}(\widehat{\rv} - \widehat{\dv}) .
\end{align}
Therefore,
\begin{equation}
\uv_{\mathrm{NDO}} = \uv_{\mathrm{INDI}} + \Am(\xv)^{-1}\Delta,
\qquad
\Delta := \widehat{\rv} - \widehat{\dv}.
\label{eq:continuous_gap}
\end{equation}

Using \eqref{eq:io},
\begin{align}
\widehat{\rv}
&= \dot{\yv}_f - \av(\xv) - \Am(\xv)\uv_f \nonumber \\
&= (\dot{\yv} + \varepsilon_f) - \av(\xv) - \Am(\xv)\uv_f \nonumber \\
&= \av(\xv) + \Am(\xv)\uv + \dv + \varepsilon_f - \av(\xv) - \Am(\xv)\uv_f \nonumber \\
&= \dv + \varepsilon_f + \Am(\xv)(\uv - \uv_f).
\end{align}
Since $\uv \approx \uv_f$, it follows that
\begin{equation}
\widehat{\rv} = \dv + \varepsilon_f .
\end{equation}
Assuming sufficiently accurate disturbance estimation, i.e., $\widehat{\dv} = \dv + \varepsilon_d$ with small estimation error $\varepsilon_d$, the gap becomes $
\Delta = \varepsilon_f - \varepsilon_d$.
Hence,
\begin{equation}
\uv_{\mathrm{NDO}} = \uv_{\mathrm{INDI}} + \Am(\xv)^{-1}(\varepsilon_f - \varepsilon_d).
\label{eq:continuous_final}
\end{equation}
Therefore, when the disturbance contribution is consistently represented in both formulations and filtering and disturbance-estimation errors are negligible, the control inputs generated by NDI+NDO and INDI  coincide.  This provides an analytical baseline for comparison, since practical implementations differ due to sampling, filtering, model uncertainty, and sensing effects.

\section{System Modeling and Control Implementation}
\label{sect:sys_modeling}
\subsection{Multirotor Dynamics and Allocation}
\label{sec:modeling}
We adopt an inertial/world frame $\mathcal{F}_W=\{O_W,\boldsymbol{e}_1,\boldsymbol{e}_2,\boldsymbol{e}_3\}$
(with $\boldsymbol{e}_3$ aligned with gravity) and a body-fixed frame
$\mathcal{F}_B=\{O_B,\boldsymbol{b}_1,\boldsymbol{b}_2,\boldsymbol{b}_3\}$ attached to the vehicle center of mass (CoM).
Let $\boldsymbol{p}\in\mathbb{R}^3$ denote the CoM position expressed in $\mathcal{F}_W$,
$\boldsymbol{v}=\dot{\boldsymbol{p}}\in\mathbb{R}^3$ the linear velocity, $\Rm\in SO(3)$ the rotation matrix from
$\mathcal{F}_B$ to $\mathcal{F}_W$, and $\Omegav\in\mathbb{R}^3$ the angular velocity expressed in
$\mathcal{F}_B$.
The state is $\xv = (\boldsymbol{p},\boldsymbol{v},\Rm,\Omegav)$.

The translational and rotational dynamics are modeled via Newton--Euler equations:
\begin{align}
m\dot{\boldsymbol{v}} &= -m g \boldsymbol{e}_3 + \Rm  \boldsymbol{F}_b + \boldsymbol{d}_{F}, \label{eq:trans_dyn}\\
\Jm\dot{\Omegav} &= -\Omegav\times (\Jm\Omegav)
+ \boldsymbol{\tau}_b + \boldsymbol{d}_{\tau}, \label{eq:rot_dyn}
\end{align}
where $m$ is the mass, $\Jm\in\mathbb{R}^{3\times 3}$ the inertia matrix expressed in $\mathcal{F}_B$,
$\boldsymbol{F}_b\in\mathbb{R}^3$ and $\boldsymbol{\tau}_b\in\mathbb{R}^3$ are the control force and torque
expressed in $\mathcal{F}_B$, and $\dv_{ext}:=[\boldsymbol{d}_{F}^\top\;\boldsymbol{d}_{\tau}^\top]^\top$ represent external
disturbances (e.g., wind). 

Let $\uv\in\mathbb{R}^{n_u}$ collect the actuator inputs. In this work,
$n_u=6$ and $u_i$ corresponds to the signed squared rotor speed of the $i$-th propeller. 
The generated body wrench is defined as
$\wv_u(\uv) :=
\begin{bmatrix}
\boldsymbol{F}_b(\uv)\\
\boldsymbol{\tau}_b(\uv)
\end{bmatrix}
\in\mathbb{R}^6.$
The total body force and torque are obtained as
$\boldsymbol{F}_b(\uv)=\sum_{i=1}^{N} \fv_i$, and $  \boldsymbol{\tau}_b = \sum_{i=1}^{N} \mv_i,$
where $\fv_i$ and $ \mv_i$ denote the force and moment contributions of the $i$-th actuation unit (AU) with respect to the CoM. They are given by
\begin{equation}
    \begin{aligned}
        \fv_i &= \fv_{p_i},\qquad
        \mv_i = \mv_{p_i} + \pv_i \times \fv_{p_i},
    \end{aligned}
\end{equation}
where $\fv_{p_i} \in \mathbb{R}^3$ is the thrust force generated by the rotation of the $i$-th propeller and $\mv_{p_i} \in \mathbb{R}^3$ is the aerodynamic moment. In particular,  $\fv_{p_i}=c_{f_i}|w_i|w_i\vv_i,$ and $
\mv_{p_i}=k_ic_{\tau_i}|w_i|w_i\vv_i$,
where $c_{f_i}\in \mathbb{R}_{\ge 0}$ and $c_{\tau_i}\in \mathbb{R}_{\ge 0}$ are positive coefficients that depend on the aerodynamic characteristics of the propeller, $\omega_i \in \mathbb{R}$ is the propeller angular speed, $k_i \in \{-1,1\}$ accounts for the propeller rotation direction, and $\vv_i \in \mathbb{S}^2$ is the unit thrust direction expressed in $\mathcal{F}_B$. The thrust direction can be parametrized by two tilting angles $\alpha_i$ and $\beta_i$ defined with respect to the position vector $\pv_i$ of the $i$-th AU.
In particular, $\alpha_i$ denotes the radial tilting angle, while $\beta_i$ denotes the tangential tilting angle.
The resulting thrust direction can be expressed as $\vv_i = \mathbf{R}_2(\beta_i)\mathbf{R}_1(\alpha_i)\ev_3$, where $\mathbf{R}_1(\alpha_i)$ and  $\mathbf{R}_2(\beta_i)$ 
denote elementary rotation matrices defining the AU orientation.
For a generic multi-rotor platform, the mapping between $\uv$ and the wrench can be expressed as  $
\wv_u(\uv) = \Fm\,\uv,$ with $
\Fm :=
\begin{bmatrix}
\Fm_1\\
\Fm_2
\end{bmatrix},
$
where $\Fm_1\in\mathbb{R}^{3\times n_u}$ maps inputs to body force and $\Fm_2\in\mathbb{R}^{3\times n_u}$
maps inputs to body torque.
The rank of $\Fm$ characterizes the actuation properties: if $\mathrm{rank}(\Fm)=6$ the platform can generate
an arbitrary wrench (full actuation), while $\mathrm{rank}(\Fm)<6$ implies under-actuation.
To evaluate disturbance rejection, wind is modeled as an aerodynamic drag-like force
\begin{equation}
\boldsymbol{d}_{\mathrm{F}} = -\Dm\big(\boldsymbol{v}-\boldsymbol{v}_{\mathrm{wind}}\big),\qquad \Dm>0,
\label{eq:wind_model}
\end{equation}
with $\boldsymbol{v}_{\mathrm{wind}}$ a stochastic wind velocity process and a damping matrix $\Dm$  containing the drag coefficients for each axis. The corresponding aerodynamic moment $\boldsymbol{\tau}_{\mathrm{ext}}$ is derived from $\boldsymbol{d}_{F}$, acting at a distance from the vehicle’s center of mass.
The multirotor acceleration dynamics can be written in the control-affine
form introduced in Section \ref{sec:equivalence}. By stacking translational and rotational
accelerations as $
\dot{\boldsymbol{y}} :=[
\ddot{\boldsymbol{p}}^\top\;
\dot{\Omegav}^\top]^\top
\in\mathbb{R}^6,$
and using \eqref{eq:trans_dyn}–\eqref{eq:rot_dyn}, we obtain
\begin{equation}
\dot{\boldsymbol{y}}
=
\underbrace{
\begin{bmatrix}
- g\boldsymbol{e}_3 \\
-\Jm^{-1}\big(\Omegav\times (\Jm\Omegav)\big)
\end{bmatrix}
}_{\boldsymbol{a}(\xv)}
+
\underbrace{
\begin{bmatrix}
\frac{1}{m}\Rm \Fm_1 \\
\Jm^{-1}\Fm_2
\end{bmatrix}
}_{\Am(\xv)}
\uv
+
\underbrace{
\begin{bmatrix}
\frac{1}{m}\boldsymbol{F}_{\mathrm{ext}} \\
\Jm^{-1}\boldsymbol{\tau}_{\mathrm{ext}}
\end{bmatrix}
}_{\boldsymbol{d}}.
\label{eq:affine_multirotor}
\end{equation}
Equation~\eqref{eq:affine_multirotor} is a specific realization of the
abstract input–output model used in Section \ref{sec:equivalence}.
Given a virtual acceleration command $
\nuv :=[
\boldsymbol{a}_c^\top\; 
\boldsymbol{\alpha}_c^\top]^\top,$
the two inversion schemes derived in Section~\ref{sec:equivalence} are
defined for the multirotor platform as follows.

\paragraph{INDI}
From the model in \eqref{eq:affine_multirotor} the incremental inversion law reads
\begin{equation}
    \uv=\uv_f+\Am(\xv)^{-1}\big(\nuv-\dot{\yv}_f \big),
\end{equation}
where $\dot{\yv}_f$ denotes the filtered acceleration measurements.

\paragraph{NDI+NDO}
The disturbance-compensated inversion law is given by 
\begin{equation}
    \uv=\Am(\xv)^{-1}\big(\nuv - \av(\xv)-\widehat{\boldsymbol{d}}\big),
    \label{eq:indi-plus-ndo}
\end{equation}
where $\widehat{\boldsymbol{d}}$ is generated by the nonlinear disturbance observer, whose dynamic is described in \eqref{eq:ob_dyn}.

\subsection{Tracking Objective and Virtual Commands}
The control objective is to track a reference position $\boldsymbol{p}_r(t)$ and reference attitude $\Rm_r(t)$.
Define the position and velocity errors as $
\boldsymbol{e}_p := \boldsymbol{p}-\boldsymbol{p}_r$, and $
\boldsymbol{e}_v := \boldsymbol{v}-\dot{\boldsymbol{p}}_r$.
For the rotational dynamics, we use the rotation error and angular-velocity error $
\boldsymbol{e}_R := \tfrac{1}{2}\big(\Rm_r^\top \Rm - \Rm^\top \Rm_r\big)^\vee$, and $
\boldsymbol{e}_{\Omegav} := \Omegav - \Rm^\top \Rm_r \Omegav_r$,
where $(\cdot)^\vee$ maps a skew-symmetric matrix to $\mathbb{R}^3$.
Virtual translational and angular acceleration commands are then defined as
\begin{align}
\boldsymbol{a}_c &:= \ddot{\boldsymbol{p}}_r - \Km_v \boldsymbol{e}_v - \Km_p \boldsymbol{e}_p,
\label{eq:ac}\\
\boldsymbol{\alpha}_c &:= \dot{\Omegav}_r - \Km_{\Omegav} \boldsymbol{e}_{\Omegav} - \Km_R \boldsymbol{e}_R,
\label{eq:alphac}
\end{align}
where $\Km_p,\Km_v,\Km_R,\Km_{\Omegav} \in \mathbb{R}^{3\times 3}$ are positive definite gain matrices.

\subsection{INDI vs. NDI+NDO Implementation}
To ensure a fair comparison between incremental and model-based inversion, both controllers employ the same virtual acceleration commands $(\boldsymbol{a}_c,\boldsymbol{\alpha}_c)$ defined in~\eqref{eq:ac}-\eqref{eq:alphac}, and operate at the same sampling rate. The two control architectures differ only in how these virtual commands are mapped into actuator inputs.
\paragraph*{INDI.}

Using the affine representation introduced in Section~\ref{sec:equivalence}, the incremental control law is implemented in discrete time as
\begin{equation}
\uv_k = \uv_{k-1} + \Am(\xv_k)^{-1}\!\Big(\nuv_k - \dot{\yv}_{f,k-1}\Big),
\label{eq:indi_update_c}
\end{equation}
where $\dot{\yv}_{f,k-1}$ is obtained from filtered acceleration measurements. In particular, the corrupted signal $\tilde{\dot{\yv}}$ is filtered using a first-order discrete low-pass filter, which attenuates high-frequency noise components
\begin{equation}
    G(z) = \frac{1 - e^{-2 \pi f_c T_s}}{1 - e^{-2 \pi f_c T_s} z^{-1}},
\end{equation}
where $f_c$ denotes the cutoff frequency (30 or 60 Hz depending on the signal) and $T_s$ is the control sampling period.
After filtering, the signals are processed through a discrete $z^{-1}$ operator (unit delay) to obtain the velocity and angular-rate derivatives used by the controller.
The update \eqref{eq:indi_update_c} avoids explicit evaluation of the full nonlinear drift terms at each step,
and thus partially compensates model mismatch and disturbances through acceleration feedback, as commonly exploited in INDI-based
flight control.
An important point is that the incremental update is driven by the difference between commanded and measured accelerations. Therefore, any slowly varying model mismatch or external disturbance that manifests in the measured acceleration is implicitly compensated in the next control update. This mechanism contributes to the reduced sensitivity of INDI to parametric uncertainties compared to full model cancellation of model-based approaches.

\paragraph*{NDI with Nonlinear Disturbance Observer (NDI+NDO).}
To mitigate the sensitivity of pure model-based inversion to parametric uncertainties and external disturbances, the inversion law is augmented with an explicit estimate of a lumped disturbance term.

Defining residual signals from measured accelerations
\begin{equation}
\boldsymbol{r}_F := m\dot{\boldsymbol{v}}-\big(-mg\,\boldsymbol{e}_3 + \Rm\boldsymbol{F}_b\big),
\label{eq:dist_residual}
\end{equation}
\begin{equation}
    \boldsymbol{r}_\tau := \Jm\dot{\Omegav} -\big(-\Omegav\times(J\Omegav) + \boldsymbol{\tau}_b\big),
\end{equation}
first-order disturbance observers are introduced
\begin{align}
    \dot{\widehat{\boldsymbol{d}}}_F &= \lambda_F\big(\boldsymbol{r}_F - \widehat{\boldsymbol{d}}_F\big),
\qquad \lambda_F>0,\\
\dot{\widehat{\boldsymbol{d}}}_\tau &= \lambda_\tau\big(\boldsymbol{r}_\tau - \widehat{\boldsymbol{d}}_\tau\big),
\qquad \lambda_\tau>0,
\end{align}
where $\lambda_F$ and $\lambda_\tau$ determine the observer bandwidth.
In compact affine form, this correspond to the disturbance-compensated inversion law~\eqref{eq:indi-plus-ndo} derived in Section~\ref{sec:equivalence}.
In discrete time, the implemented control law is: 
\begin{equation}
\uv_k=\Am(\xv_k)^{-1}\big(\nuv_k-\av(\xv_k)-\widehat{\dv}_k\big),
\end{equation}
where $\widehat{\dv}_k$ is obtained from the discretized
observer dynamics.
\begin{rem}
    In contrast to INDI, which compensates the disturbances implicitly through incremental acceleration feedback, NDI+NDO relies on explicit disturbance-estimation dynamics. The observer gains ($\lambda_F,\lambda_\tau$) determine a trade-off between disturbance-rejection speed and noise sensitivity, thereby introducing an additional tuning that does not appear in the pure inversion scheme.
\end{rem}

\begin{table}[t]
\centering
\caption{Nominal Aerial Robot Platform Parameters.}
\label{tab:minithex_params}
\begin{tabular}{lll}
\hline
\bf Parameter & \bf Symbol & \bf Value \\
\hline
Mass & $m_n$ & $0.935~\mathrm{kg}$ \\
Inertia matrix & $J_n$ & $\mathrm{diag}(1.49,\,1.71,\,2.77)\times 10^{-3}~\mathrm{kg\,m^2}$ \\
Number of rotors & $N$ & $6$ \\
Arm length & $l$ & $0.155~\mathrm{m}$ \\
Thrust coefficient & $c_f$ & $6.7\times 10^{-5}$ \\
Torque coefficient & $c_t$ & $1.0\times 10^{-6}$ \\
Rotor tilt angles & $\alpha, \beta$ & $26^\circ,\;14^\circ$ \\
\hline
\end{tabular}
\end{table}
\section{Robustness Evaluation Framework}
\label{sec:robustness_framework}

The robustness of the proposed control architectures is assessed through a structured simulation campaign
designed to evaluate their sensitivity to parametric uncertainties, external disturbances, and measurement noise.
Both NDI+NDO and INDI controllers are tested under identical conditions, gains and sampling settings.
\subsection{Reference trajectory}
To evaluate tracking performance under dynamically persistent excitation, a planar Lissajous reference trajectory is adopted. The trajectory is defined as \mbox{$
        \xv_r(t) = A_x \sin(a_x t + \delta_x)$,}
        $\yv_r(t) = A_y \sin(a_y t)$, and 
        $\zv_r(t) = z_0$,
resulting in a planar Lissajous motion in the $xy$-plane while maintaining constant the altitude. The parameters used in simulation are $A_x=1.0~\mathrm{m}$, $A_y=0.7~\mathrm{m}$, $a_x=1~\mathrm{rad/s}$, $a_y=2~\mathrm{rad/s}$, $\delta_x=\frac{\pi}{3}$, and $z_0=0.8~\mathrm{m}$. Since only translational dynamics are excited, the reference attitude is kept constant, i.e., \mbox{$\Rm_r = \mathbf{I}_3$} and $\Omegav_r = \boldsymbol{0}$.

\subsection{Simulation and Controller Settings}
All simulations are performed in discrete time with a sampling period $T_s=10^{-4}$~s and total duration $T=10$~s. The initial conditions are set to $\boldsymbol{p}_0=[1\; 0.2\; 0.1]^\top$~m, $\boldsymbol{v}_0=\mathbf{0}$, $\Omegav_0=\mathbf{0}$, and $\mathbf{R}_0=\mathbf{I}_3$.
The same outer-loop tracking gains are used for both the controllers and are set as
       $\Km_p=\mathrm{diag}(15.5, \; 35,\; 16.7)$, $
    \Km_v=\mathrm{diag}(9.5, \; 38,\; 13.5)$, 
    $\Km_R=\mathrm{diag}(3.2, \; 3.5,\; 6.6)$,and $
    \Km_\omega=\mathrm{diag}(14, \; 15,\; 26)$.
For the NDI+NDO architecture, the observer gains are selected as $\lambda_F=30$ for the translational branch and $\lambda_\tau=60$ for the rotational branch. The observer gains $\lambda_F$ and $\lambda_\tau$ were selected to achieve comparable effective bandwidth to the acceleration filtering used in INDI. Specifically, the observer poles were placed near the cutoff frequencies of the corresponding INDI low-pass filters.
In table~\ref{tab:minithex_params}, the nominal parameters of the platform are shown. 
\subsection{Uncertainty, Disturbance, and Noise Modeling}
To assess robustness under realistic operating conditions, three sources of non-idealities are considered: (i) parametric uncertainty, (ii) external wind disturbances, and (iii) measurement noise. Their modeling is described below. 
\paragraph*{Parametric uncertainty.}
Parametric robustness is evaluated by varying mass and inertia around their nominal values $(m_n, \Jm_n)$. The parameters are modeled as $
    m=m_n(1+\Delta m)$, and $\Jm=\Jm_n(1+\Delta J)$,
where $\Delta m$ and $\Delta J$ represent relative deviations from the nominal parameters. In the representative scenarios considered in this work, parametric uncertainty is introduced only in selected cases. When present, relative deviations of $\pm20\%$ are applied to both mass and inertia.
Both INDI and NDI+NDO compute the matrix $\Am(\xv)$ using the same nominal parameters $(m_n, \Jm_n)$. The true plant dynamics, however, evolve according to the perturbed parameters. This ensures a fair comparison, as both architectures are subject to identical model mismatch in the inversion process.

\begin{figure}[t]
    \centering
    \includegraphics[width=1\linewidth]{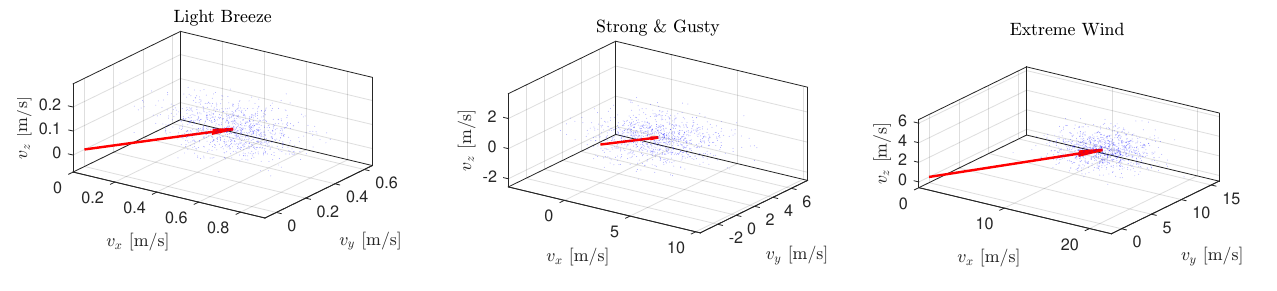}
    \caption{Wind profiles considered in the robustness simulations. The red vector indicates the mean wind component, while the blue dots represent the variability around this value. }
    \label{fig:kind_of_wind}
\end{figure}

\paragraph*{Wind disturbances.}
External disturbances are modeled according to \eqref{eq:wind_model}\footnote{While we adopt a stochastic drag-based model for this study, future work will involve testing under more complex aerodynamic models such as the Dryden wind turbulence model.}.
A set of wind models with increasing intensity and variability is defined, ranging from calm conditions to severe turbulence.
Each wind model is parameterized by a mean wind vector $\boldsymbol{\mu}$, a covariance matrix $\Sigma$,
and an aerodynamic damping matrix $\Dm$. The numerical values used in simulation are reported in
Table~\ref{tab:wind_scenarios} and visually summarized in Fig.~\ref{fig:kind_of_wind}.

\begin{table}[t]
\centering
\footnotesize
\setlength{\tabcolsep}{7pt}
\renewcommand{\arraystretch}{1.0}
\caption{Parameters used to generate the different wind scenarios.}
\begin{tabular}{c c c c}
\hline
\begin{tabular}{c}
\textbf{Wind}\\\textbf{Scenario}
\end{tabular}
& $\boldsymbol{\mu}$ 
& $\Sigma$ 
& $\Dm$ \\
\hline

No Wind 
& $\boldsymbol{0}$ 
& $\boldsymbol{0}$ 
& $\boldsymbol{0}$ \\

Light Breeze 
& $[0.5\;0.3\;0.1]^\top$
& $\mathrm{diag}(0.15^2,\,0.10^2,\,0.05^2)$
& $2\bm{I}_3$ \\

Strong \& Gusty
& $[3.0\;2.0\;0.5]^\top$
& $\begin{bmatrix}
4.0 & 0.5 & 0.1\\
0.5 & 2.0 & 0.0\\
0.1 & 0.0 & 0.5
\end{bmatrix}$
& $2\bm{I}_3$ \\

Extreme Wind
& $[14.0\;9.0\;3.0]^\top$
& $\begin{bmatrix}
6.0 & 0.7 & 0.2\\
0.7 & 3.0 & 0.0\\
0.2 & 0.0 & 0.7
\end{bmatrix}$
& $3\bm{I}_3$ \\

\hline
\end{tabular}
\label{tab:wind_scenarios}
\end{table}

\paragraph*{Measurement noise.}
Measurement noise is injected in the acceleration and angular velocity channels according to
\begin{equation}
    \tilde{\yv}=\yv+\boldsymbol{n}, \qquad \boldsymbol{n}\sim \mathcal{N}(\boldsymbol{\mu},\Sigma),
\end{equation}
where $\boldsymbol{\mu}$ represents the bias component and $\Sigma$ the noise covariance.
These signals directly affect the INDI update law, and indirectly influence the disturbance estimation in the NDI+NDO architecture.
Different sensor quality levels are defined through distinct values of the mean and the covariance parameters, as reported in Table~\ref{tab:quality_sensors}.
The Ideal case corresponds to noise-free measurements and is used in the nominal scenario.
The "Low Quality" sensor model intentionally represents degraded or poorly calibrated sensing conditions rather than modern high-grade IMUs. The bias and variance levels were selected to stress the disturbance rejection mechanisms and evaluate robustness margins under adverse sensing conditions. The "High Quality" configuration reflects specifications consistent with commercially available MEMS IMUs used in aerial robotics platforms.

\begin{table}[h!]
    \centering
    \caption{{Accelerometer and gyroscope parameters used to simulate different levels of sensor quality.}}
    \renewcommand{\arraystretch}{1.2}
    \setlength{\tabcolsep}{4pt}
    \begin{tabular}{c|cccc}
    \hline
        \textbf{Sensor Quality} & \textbf{Bias acc.}  & \textbf{Var. acc.} & \textbf{Bias gyro.} & \textbf{Var. gyro.}  \\
         & [$m/s^2$] & [$(m/s^2)^2$] & [$rad/s$] & [$(rad/s)^2$]\\ 
         \hline 
        Ideal & 0.0 & 0.0 & 0.0 & 0.0 \\
        High Quality & 0.01 & 0.001 & 0.001 & 0.001\\
       
        Low Quality & 1.0 &0.005 & 0.05 & 0.005 \\
        
        \hline
    \end{tabular}
    
    \label{tab:quality_sensors}
\end{table}

\subsection{Performance Metrics}
The comparative analysis between INDI and NDI+NDO is carried out using tracking-performance metrics and control-effort discrepancy metrics.
\paragraph*{Full-horizon RMS.}
Tracking performance is evaluated through the Root Mean Square (RMS) error on position and orientation over the full simulation horizon $T$.
The position RMS is defined as $\mathrm{RMS}_p =
\sqrt{
\frac{1}{T}
\int_0^T
\|\boldsymbol{e}_p(t)\|^2 dt
},
$
and the orientation RMS as
$
\mathrm{RMS}_R =
\sqrt{
\frac{1}{T}
\int_0^T
\|\boldsymbol{e}_R(t)\|^2 dt
}.$

These metrics quantify the overall tracking performance, including both transient and steady-state behavior.
\paragraph*{Steady-state RMS.}
In addition to full-horizon performance, a steady-state RMS is computed over the interval $t \in [ t_{ss}, T]$, with $t_{ss}=2.5~\mathrm{s}$ fixed for all simulations. The steady-state RMS is defined as
\begin{equation}
    \mathrm{RMS}^{ss}_{k}=\sqrt{\frac{1}{T-t_{ss}}\int_{t_{ss}}^T\|\boldsymbol{e}_{k}(t)\|^2 dt },\qquad k=p,R.
\label{eq:rms_ss}
\end{equation}
This metric isolates the residual tracking error after the initial transient.
\paragraph*{Control wrench RMS.}
To quantify the difference between the control actions generated by the two architectures, the RMS error of the wrench discrepancy is considered and defined as
\begin{equation}
    \mathrm{RMS}_{w_u} =
\sqrt{
\frac{1}{T}
\int_0^T
\|{{\wv_u}}_{\mathrm{INDI}}(t)- {{\wv_u}}_{\mathrm{NDI+NDO}}(t) \|^2 \mathrm{dt}
}.
\label{eq:rms_wrench}
\end{equation}

\paragraph*{Control energy.} In addition to the wrench discrepancy metric in~\eqref{eq:rms_wrench}, 
we quantify the overall actuation effort through the control energy $
E_{\wv_u} = \int_{0}^{T} \| \wv_u(t) \|^2 \, \mathrm{dt},
\label{eq:control_energy}$
where $\wv_u(t)\in\mathbb{R}^6$ denotes the commanded body wrench. 
This metric highlights differences in actuator usage that may not be visible in tracking RMS, 
and it is relevant for practical considerations such as saturation margins and power/thermal loading.

\begin{table}[t]
\renewcommand{\arraystretch}{1.4}
\centering
\caption{Representative scenarios used for the comparative evaluation of INDI and NDI+NDO under realistic operating conditions.}
\label{tab:six_case}
\resizebox{\columnwidth}{!}{%
\begin{tabular}{c|cccc}
\hline
\textit{Case} & Mass Var. & Inertia Var. & Wind & Sensor \\ \hline
Nominal & $0\%$ & $0\%$ & No wind & Ideal \\
Light Model & $-20\%$ & $-20\%$ & Light Breeze & High Quality \\
Heavy Model & $+20\%$ & $+20\%$ & Light Breeze & High Quality \\
Bad Sensors & $0\%$ & $0\%$ & Light Breeze & Low Quality \\
Extreme Wind & $0\%$ & $0\%$ & Extreme Wind & High Quality \\
Combined Stress & $+20\%$ & $+20\%$ & Strong \& Gusty & Low Quality \\
\hline
\end{tabular}%
}
\end{table}

\subsection{Comparative Evaluation Scenarios}
The performance of INDI and NDI+NDO is assessed through six representative operating scenarios combining different levels of parametric uncertainty, wind disturbances, and measurement noise. All simulations use the same reference trajectory and identical controller gains to ensure a fair comparison. The scenarios considered are reported in Table~\ref{tab:six_case}. For scenarios including stochastic measurement noise, a Monte Carlo campaign with $N_{\mathrm{rep}}=20$ independent noise realizations is performed. Increasing the number of realizations did not alter the qualitative trends observed.  For each configuration, mean RMS values are reported.

\section{Comparative Results}
\label{sec:discussion}
This section presents the comparative simulation results obtained by applying INDI and NDI+NDO to the fully-actuated multirotor platform described in Section~\ref{sec:modeling}, under the representative scenarios defined in Section~\ref{sec:robustness_framework}. The objective is to assess how the structural equivalence discussed in Section~\ref{sec:equivalence} translates into performance under progressively increasing non-ideal conditions. The analysis is structured as follows. First, qualitative trajectory tracking results are discussed. Then, full-horizon and steady-state RMS metrics are examined to quantify tracking accuracy. Finally, the RMS difference of the generated control wrench is analyzed to highlight the practical difference between the two control architectures. 
\subsection{Qualitative Trajectory Comparison}
\begin{figure}[t!]
    \centering
\includegraphics[width=1.03\linewidth]{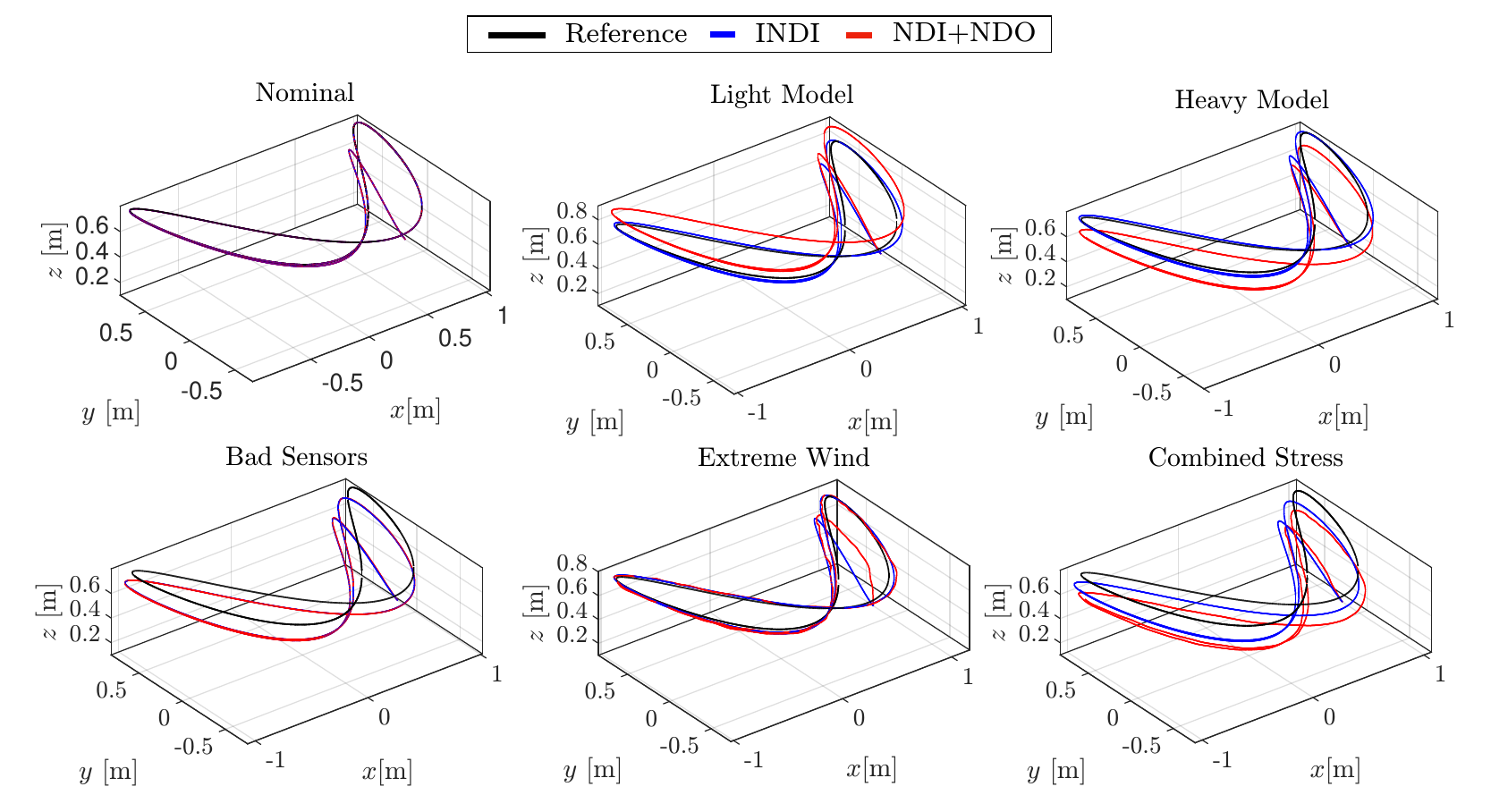}
    \caption{Three-dimensional tracking performance of INDI (blue) and NDI+NDO (red) for the six representative scenarios. The reference trajectory is shown in black. }
    \label{fig:traj_comparison}
\end{figure}
Figure~\ref{fig:traj_comparison} illustrates the three-dimensional trajectories obtained with INDI and NDI+NDO across the six representative scenarios.
\paragraph{Nominal (Ideal) Conditions} INDI and NDI+NDO produce identical trajectories and control wrenches. The RMS difference between control actions is on the order of $10^{-14}$ confirming exact practical equivalence within numerical precision.
\paragraph{Non-ideal Conditions}
When non-ideal effects are introduced, visible differences begin to emerge. In the Light and Heavy Model scenarios, NDI+NDO exhibits evident deviations from the reference, whereas INDI maintains a trajectory closer to the desired path. Interestingly, the deviation observed for NDI+NDO in the Light Model case appears approximately mirrored in the Heavy Model scenario. This suggests that NDI+NDO reacts differently to negative and positive parameter variations. In the Bad Sensor and Extreme Wind scenarios, both controllers have comparable degradation. The most evident difference appears in the Combined Stress case, where NDI+NDO displays larger deviations while INDI preserves a trajectory closer to  the nominal behavior. These qualitative observations highlight divergences between the two control architectures that depend on the considered operating scenario. These differences are quantified in the following RMS-based analysis.

\subsection{Full-Horizon RMS Analysis}
\begin{figure}[t]
    \centering
    \includegraphics[width=1.05\linewidth]{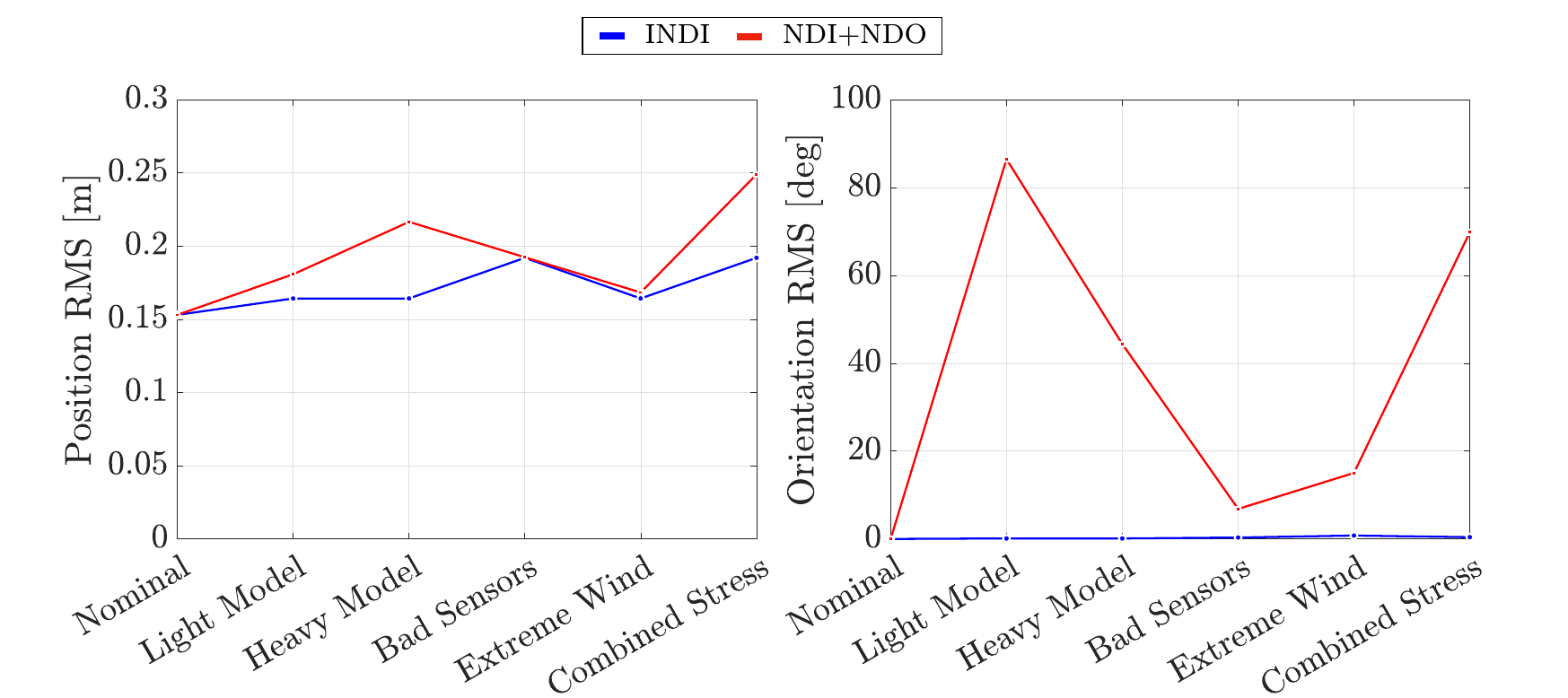}
    \caption{Full-horizon RMS tracking errors for INDI (blue) and NDI+NDO (red) across the six representative scenarios. The figure reports position (left) and orientation (right) RMS values, computed over the entire simulation horizon.}
    \label{fig:fullHorizon_RMS}
\end{figure}

Figure~\ref{fig:fullHorizon_RMS} reports the position and orientation RMS errors computed over the entire simulation horizon for the six representative scenarios.
\paragraph*{Position tracking.} Under nominal conditions, both architectures exhibit identical position RMS values, confirming practical equivalence in ideal operating conditions. In scenarios combining model uncertainty with mild wind and high quality sensors (Light and Heavy Model cases), NDI+NDO exhibits an increase in position RMS, whereas INDI remains closer to the nominal level. This suggests a stronger sensitivity of the model-based inversion mechanism to mass/inertia mismatch in the translational channel. In the Bad Sensors and Extreme Wind scenarios, both controllers maintain comparable performance, with moderate variations across the scenarios. The largest discrepancy is observed in the Combined Stress case, where simultaneous model mismatch, wind disturbance, and measurement noise lead to marked degradation for NDI+NDO, whereas INDI preserves lower RMS values.
\paragraph*{Orientation tracking.}
A pronounced difference emerges in the attitude RMS. While INDI maintains consistently low orientation errors across all scenarios, NDI+NDO exhibits a substantial increase in RMS under several non-ideal operating conditions.
Under nominal conditions, both controllers achieve negligible errors. The highest peaks are observed in the Light Model and Combined Stress cases, with a  slightly lower, yet still significant, increase in the Heavy Model case. The discrepancy between Light and Heavy Model suggests an asymmetric sensitivity with respect to the sign of the parameter variation, with negative variation leading to a larger degradation.
In particular, when inertia is underestimated, the inversion gain increases, which may amplify residual disturbance estimation errors in the rotational channel. Conversely, inertia overestimation leads to a more  conservative inversion gain that attenuates compensation but may reduce amplification of observer bias.
In contrast, when no model uncertainty is present (Nominal, Bad Sensor, and Extreme Wind cases), the orientation RMS of NDI+NDO remains significantly lower. Overall, these results highlight a strong sensitivity of NDI+NDO architecture to parameter uncertainty in combination with external disturbances and sensing effects in the rotational dynamics, whereas INDI remains largely unaffected.
\subsection{Steady-State RMS Analysis}
\begin{figure}[t]
    \centering
    \includegraphics[width=1.02\linewidth]{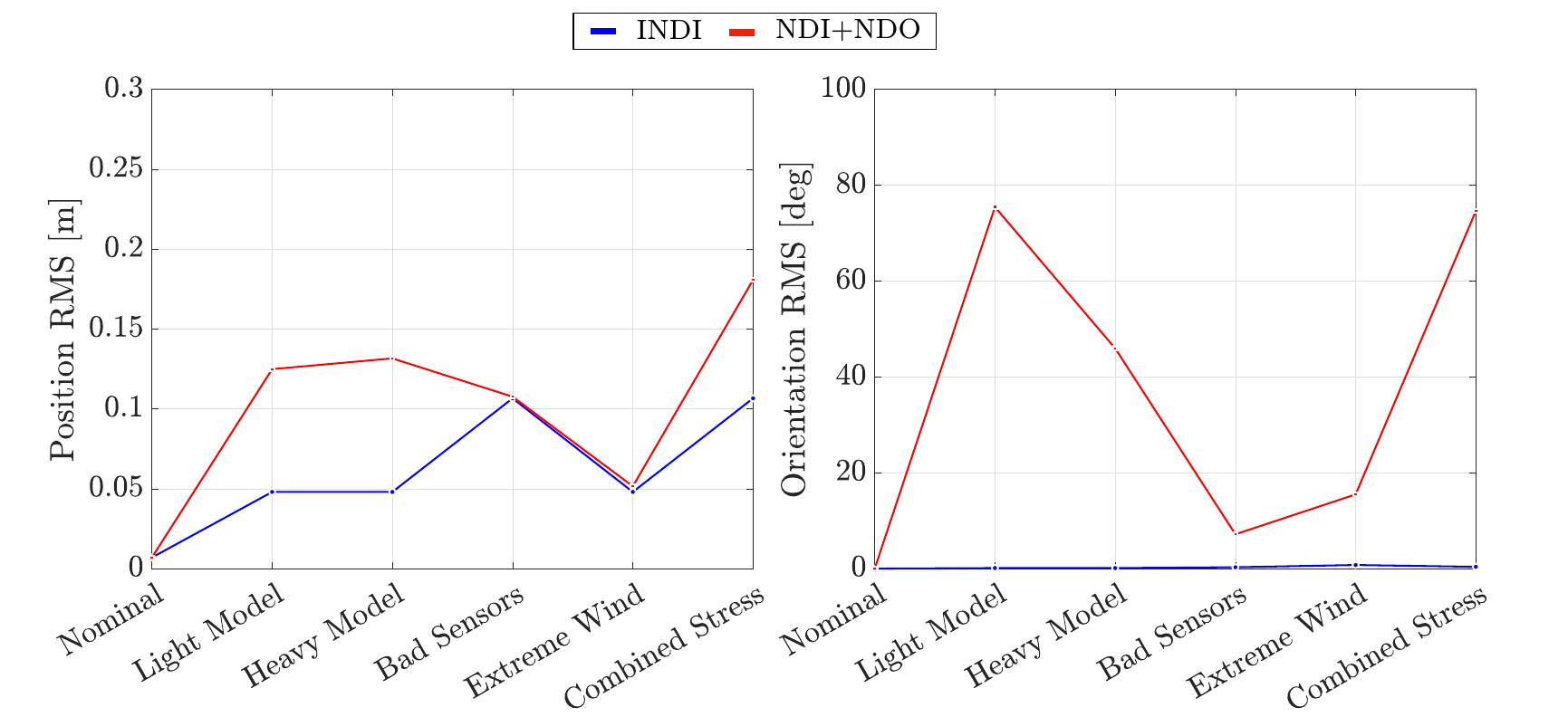}
    \caption{RMS tracking error for INDI (blue) and NDI+NDO (red) at steady-state across the six representative scenarios. The figure reports position (left) and orientation (right) RMS values, computed for $t \ge t_{ss}$.}
    \label{fig:ss_RMS}
\end{figure}
The steady-state analysis complements the full-horizon study by isolating the residual tracking error after the initial transient phase. It clarifies whether the discrepancies observed in the full-horizon RMS are primarily driven by transient effects or persist after convergence.
Figure~\ref{fig:ss_RMS} reports the steady-state RMS values computed for $t\ge t_{ss}$.
\paragraph*{Position tracking}
Under nominal conditions, both architectures achieve identical position residual errors. However, in the Light and Heavy Model cases the steady-state RMS of NDI+NDO remains significantly higher, whereas INDI maintains a residual error comparable to the nominal case. This confirms that the divergence observed in the full-horizon results is not solely due to transient dynamics, but reflects a difference in steady-state tracking. In the Bad Sensors case, both controllers exhibit a similar increase in steady-state RMS error, as in the case of Extreme Wind. The largest steady-state error is visible for the Combined Stress case. 
\paragraph*{Orientation tracking}
The steady-state orientation RMS confirms that the large discrepancies observed in the full-horizon analysis are not solely transient effects. While INDI maintains negligible residual attitude errors across all scenarios, NDI+NDO exhibits elevated steady-state RMS values. The highest error occurs in the Light Model and Combined Stress cases, with a slightly lower but still significant increase in the Heavy Model case. In contrast, when no model uncertainty is introduced (Nominal, Bad Sensor, and Extreme Wind scenarios), the steady-state orientation RMS of NDI+NDO remains lower. These results reveal that the degradation in attitude tracking for NDI+NDO persists at regime under non-ideal operating conditions, whereas INDI preserves near-zero steady-state errors throughout all the tested scenarios. 
\subsection{Control Wrench Difference Analysis}
\begin{figure}[t]
    \centering
    \includegraphics[width=1\linewidth]{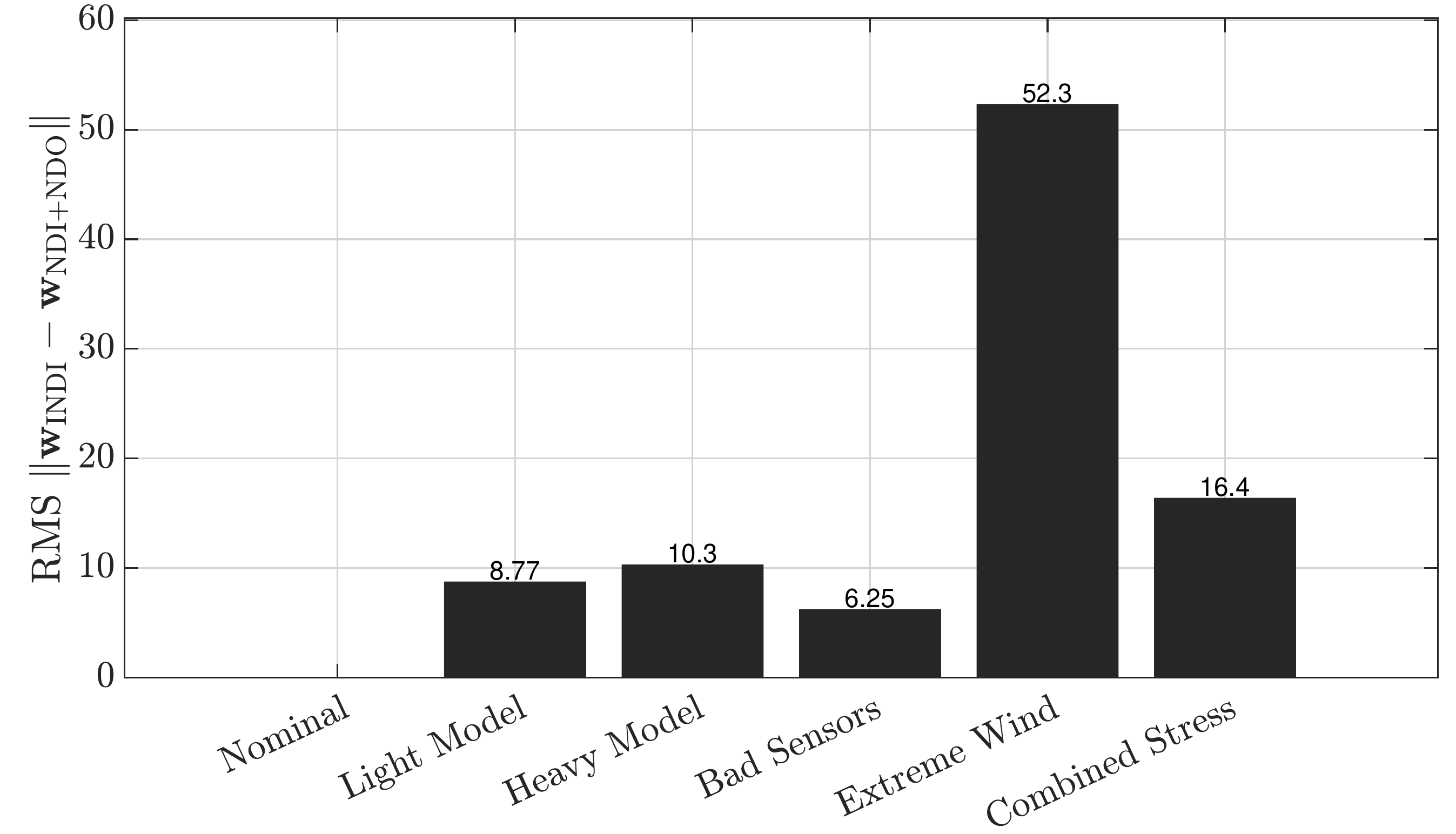}
    \caption{RMS of the difference between the control wrenches produced by INDI and NDI+NDO across the six representative scenarios. The Nominal case yields an RMS difference at machine precision ($\approx 5 \times 10^{-14}$), therefore the numerical value is omitted from the figure.}
    \label{fig:diff_wrench}
\end{figure}

Figure~\ref{fig:diff_wrench} reports the RMS of the difference between the control wrenches generated by INDI and NDI+NDO. Under nominal conditions, the RMS difference is of the order of $10^{-14}$, confirming exact equivalence in the nominal case. This result is consistent with the structural equivalence established in Section~\ref{sec:equivalence}. When non-ideal effects are introduced, the divergence between the control actions increases. Moderate discrepancies are observed in Light Model, Heavy Model, Bad Sensor and Combined Stress cases. The largest deviation occurs in the Extreme Wind scenario, indicating that the two architectures respond differently to strong external disturbances. Despite these differences in control wrenches, tracking performance does not always degrade proportionally, as shown in the previous subsections. This indicates that the two control architectures can yield comparable tracking behaviour while generating significantly different control wrench commands. 
\subsection{Wrench Energy Comparison}
\begin{figure}[t!]
    \centering
\includegraphics[width=0.99\linewidth]{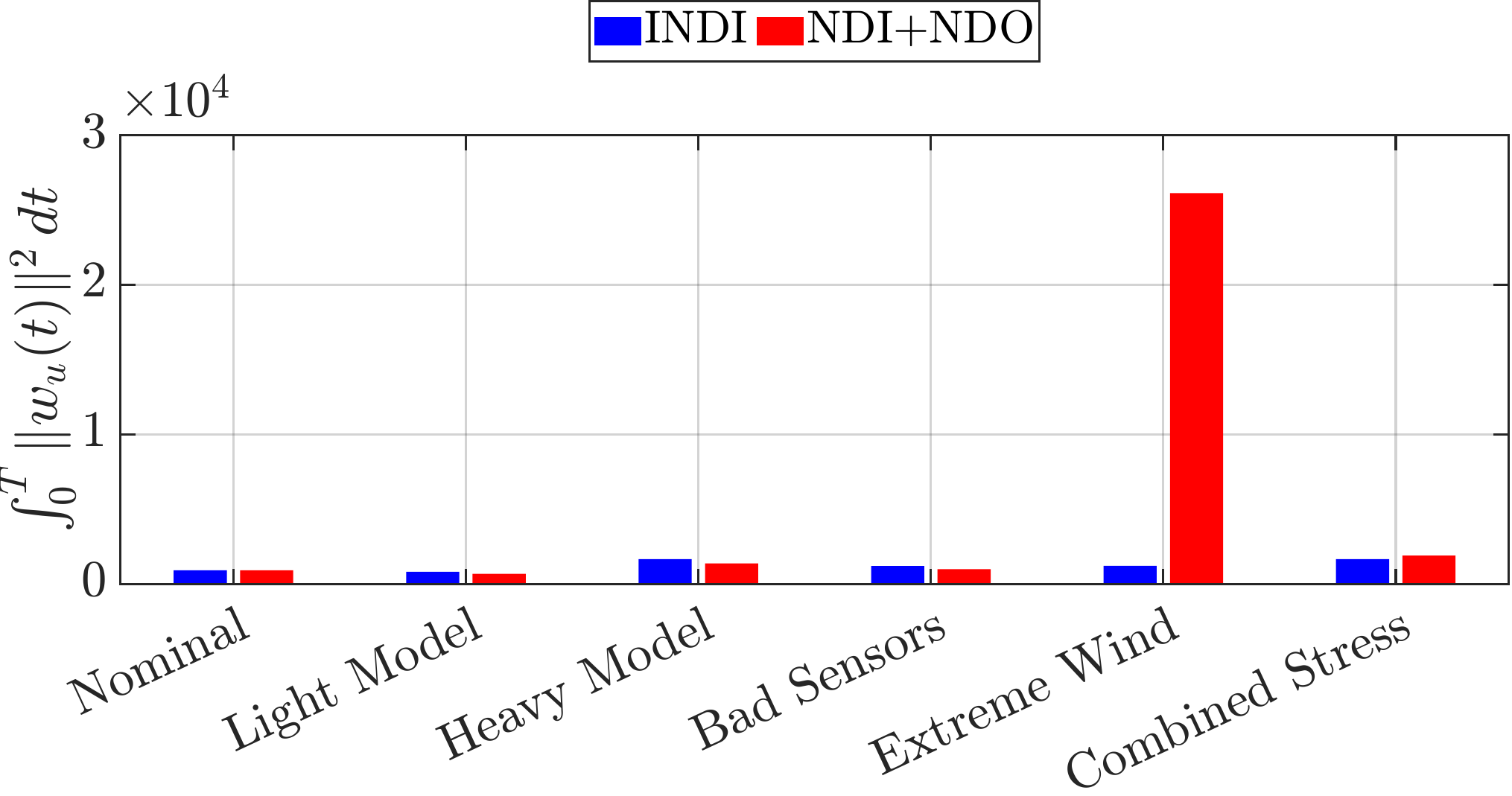}
    \caption{Energy of the commanded wrench, computed as $\int_0^T\|\wv_u(t)\|^2 dt$, both for INDI (blue) and NDI+NDO (red) across all the representative scenarios.}
    \label{fig:energy_wrench_comparison}
\end{figure}

Figure~\ref{fig:energy_wrench_comparison} reports the energy generated by INDI and NDI+NDO across the six representative scenarios.
Under nominal conditions, both the control architectures exhibit the same energy values. In non-ideal scenarios, differences in energy emerge. In particular, in the Extreme Wind case, NDI+NDO requires a significantly higher control energy, whereas INDI maintains a comparatively moderate increase. This behavior indicates that the two disturbance compensation mechanisms react differently to strong external perturbations in terms of control energy demand.
\subsection{Discussion}
The analysis reveals a clear difference between ideal and non-ideal operating conditions. Under nominal assumptions, INDI and NDI+NDO produce identical tracking performance and control actions, confirming the structural equivalence established in Section~\ref{sec:equivalence}. When non-ideal effects are introduced, some differences emerge. While both the control architectures remain stable across all the tested scenarios, NDI+NDO exhibits a stronger sensitivity to model uncertainty, in particular in the rotational dynamics.
The steady-state results confirm that the divergence between the two control architectures cannot be attributed solely to initial transient dynamics. Even after convergence, NDI+NDO retains larger errors under non-ideal operating conditions, whereas INDI maintains near nominal performance.
The control wrench analysis further shows that similar tracking does not necessarily imply similar control wrench commands.
The energy analysis further indicates that structural equivalence under ideal assumptions
does not necessarily imply identical actuator effort in practical operating conditions. This may have relevant implications in terms of saturation margins and efficiency in 
real flight platforms.
Overall, the results suggest that structural equivalence under ideal modeling assumptions does not guarantee identical robustness properties in the presence of model uncertainties, external disturbances and measurement noise.
\section{Conclusion and Future Works}
\label{sec:concl}

This work presented a comparative robustness analysis of Incremental Nonlinear Dynamic Inversion (INDI) and Nonlinear Dynamic Inversion augmented with a nonlinear disturbance observer (NDI+NDO) for fully actuated aerial robots. An analytical comparison in nominal disturbance-free conditions showed that -- as expected for fully-actuated platforms --  the two controllers can produce identical control inputs, providing a useful baseline for comparative evaluation.

The systematic simulation campaign demonstrated that this nominal correspondence does not necessarily translate into identical practical behavior under realistic operating conditions. When model uncertainty, wind disturbances, sensing degradation, sampling, and filtering effects are introduced, meaningful differences emerge in tracking performance, robustness, and control effort. In particular, the results showed that controller performance is strongly scenario dependent: INDI demonstrated stronger robustness in several model-mismatch and combined-stress scenarios, while NDI+NDO remained competitive in nominal conditions and selected disturbance cases.  

Overall, across the considered simulation campaign, INDI consistently matched nominal NDI+NDO performance while demonstrating superior robustness under non-ideal operating conditions, particularly in the presence of model mismatch, sensing degradation, and combined stress factors. In addition, NDI+NDO generally required higher commanded wrench energy in the most demanding scenarios, suggesting increased actuation burden under severe disturbances. 

This study therefore provides practical insight into the strengths and limitations of incremental and observer-based inversion strategies for aerial robotics, supporting more informed controller selection for different operating scenarios. Future work will investigate robustness--bandwidth trade-offs, broader disturbance-observer formulations, and experimental validation on real aerial platforms.

\bibliographystyle{IEEEtran}
\bibliography{Bib/bibAlias,Bib/bibAF,Bib/bibCustom, Bib/custom_benedetta,Bib/custom}

@inProceedings{2023e-AfiCorSabAboSidFra,
author  = {A. Afifi and G. Corsini and Q. Sabl\'e and Y. Aboudorra and D. Sidobre and A. Franchi},
title   = {Physical Human-Aerial Robot Interaction and Collaboration: Exploratory Results and Lessons Learned},
booktitle = "2023 "#icuas,
year	  = {2023},
address   = {Warsaw, Poland},
month     = {June},
pages     = {},
}

@STRING{icuas	= "Int. Conf. on Unmanned Aircraft Systems"}

@STRING{sv = "Springer"}

@article{colomina2014unmanned,
  title={Unmanned aerial systems for photogrammetry and remote sensing: A review},
  author={Colomina, Ismael and Molina, Pere},
  journal={Journal of photogrammetry and remote sensing},
  volume={92},
  pages={79--97},
  year={2014},
}

@article{Isidori1995,
author="A. Isidori",
title="Elementary Theory of Nonlinear Feedback for Multi-Input Multi-Output Systems",
journal="Nonlinear Control Systems",
year="1995",
publisher=sv,
pages="219-291",
doi = {10.1007/978-1-84628-615-5}
}

@article{Sieberling2010,
author = {Sieberling, S. and Chu, Q. P. and Mulder, J. A.},
title = {Robust Flight Control Using Incremental Nonlinear Dynamic Inversion and Angular Acceleration Prediction},
journal = {Journal of Guidance, Control, and Dynamics},
volume = {33},
number = {6},
pages = {1732-1742},
year = {2010},
doi = {10.2514/1.49978},

URL = { 
    
        https://doi.org/10.2514/1.49978
    
    

},
eprint = { 
    
        https://doi.org/10.2514/1.49978
    
    

}

}

@inproceedings{acquatella2012,
  title={Robust Nonlinear Spacecraft Attitude Control using Incremental Nonlinear Dynamic Inversion.},
  author={Acquatella, Paul and Falkena, Wouter and van Kampen, Erik-Jan and Chu, Q Ping},
  booktitle={AIAA Guidance, Navigation, and Control Conference},
  pages={4623},
  year={2012}
}

@article{wang2019,
  title={Stability analysis for incremental nonlinear dynamic inversion control},
  author={Wang, Xuerui and Van Kampen, Erik-Jan and Chu, Qiping and Lu, Peng},
  journal={Journal of Guidance, Control, and Dynamics},
  volume={42},
  number={5},
  pages={1116--1129},
  year={2019},
  publisher={American Institute of Aeronautics and Astronautics}
}

@misc{liu_2022,
      title={Incremental Control System Design and Flight Tests of a Micro-Coaxial Rotor UAV}, 
      author={Z. C. Liu and Y. F. Zhang and Z. D. zhang and H. X. Chen},
      year={2022},
      eprint={2203.07087},
      archivePrefix={arXiv},
      primaryClass={eess.SY},
      url={https://arxiv.org/abs/2203.07087}, 
}

@article{Sun2021,
  title={Incremental nonlinear fault-tolerant control of a quadrotor with complete loss of two opposing rotors},
  author={Sun, Sihao and Wang, Xuerui and Chu, Qiping and de Visser, Coen},
  journal={IEEE Transactions on Robotics},
  volume={37},
  number={1},
  pages={116--130},
  year={2020},
  publisher={IEEE}
}

@article{ryll,
  title={6D interaction control with aerial robots: The flying end-effector paradigm},
  author={Ryll, Markus and Muscio, Giuseppe and Pierri, Francesco and Cataldi, Elisabetta and Antonelli, Gianluca and Caccavale, Fabrizio and Bicego, Davide and Franchi, Antonio},
  journal={The International Journal of Robotics Research},
  volume={38},
  number={9},
  pages={1045--1062},
  year={2019},
  publisher={SAGE Publications Sage UK: London, England}
}

\end{document}